\documentclass[default,iicol]{sn-jnl}

\usepackage[X2, T2A]{fontenc}
\usepackage[utf8]{inputenc}

\usepackage{graphicx}%
\usepackage{caption} 
\usepackage{array} 
\usepackage{float}
\usepackage{multirow}%
\usepackage{amsmath,amssymb,amsfonts}%
\usepackage{amsthm}%
\usepackage{mathrsfs}%
\usepackage[title]{appendix}%
\usepackage{xcolor}%
\usepackage{textcomp}%
\usepackage{manyfoot}%
\usepackage{booktabs}\let\cline\cmidrule
\usepackage{algorithm}%
\usepackage{algorithmicx}%
\usepackage{algpseudocode}%
\usepackage{listings}%
\usepackage[bulgarian,main=english]{babel}

\DeclareTextSymbolDefault{\cyryat}{X2}
\DeclareTextSymbolDefault{\cyrbyus}{X2}
\DeclareTextSymbolDefault{\cyriotbyus}{X2}

\theoremstyle{thmstyleone}%
\theoremstyle{thmstyletwo}%

\theoremstyle{thmstylethree}%

\usepackage{tikz,xcolor,hyperref}

\definecolor{lime}{HTML}{A6CE39}
\DeclareRobustCommand{\orcidicon}{%
	\begin{tikzpicture}
	\draw[lime, fill=lime] (0,0) 
	circle [radius=0.16] 
	node[white] {{\fontfamily{qag}\selectfont \tiny ID}};
	\draw[white, fill=white] (-0.0625,0.095) 
	circle [radius=0.007];
	\end{tikzpicture}
	\hspace{-2mm}
}

\foreach \x in {A, ...,Z}{%
	\expandafter\xdef\csname orcid\x\endcsname{\noexpand\href{https://orcid.org/\csname orcidauthor\x\endcsname}{\noexpand\orcidicon}}
}

\usepackage[hang, flushmargin, symbol]{footmisc}
\usepackage{etoolbox}
\makeatletter
\let\fm@makefntext\@makefntext
\patchcmd\maketitle{\if@twocolumn}{\let\@makefntext\fm@makefntext\if@twocolumn}{}
\makeatother

\begin{document}

\title[Post-OCR Text Correction for Bulgarian Historical Documents]{Post-OCR Text Correction for Bulgarian Historical Documents}


\author*[1]{\fnm{Angel} \sur{Beshirov}\orcidA{}}\email{beshirov@uni-sofia.bg}

\author[1]{\fnm{Milena} \sur{Dobreva}\orcidB{}}\email{milena.dobreva@gmail.com}

\author[1]{\fnm{Dimitar} \sur{Dimitrov}\orcidC{}}\email{mitko.bg.ss@gmail.com}

\author[2]{\fnm{Momchil} \sur{Hardalov}\orcidD{}}\email{momchilh@amazon.com}

\author[1]{\fnm{Ivan} \sur{Koychev}\orcidE{}}\email{koychev@fmi.uni-sofia.bg}

\author[3]{\fnm{Preslav} \sur{Nakov}\orcidF{}}\email{preslav.nakov@gmail.com}

\affil*[1]{\orgdiv{FMI}, \orgname{Sofia University ``St.\ Kliment Ohridski''}, \orgaddress{\city{Sofia}, \country{Bulgaria}}}

\affil[2]{\orgdiv{AWS AI Labs}, \country{Spain}}

\affil[3]{\orgname{Mohamed bin Zayed University of Artificial Intelligence}, \orgaddress{\country{UAE}}}

\abstract{The digitization of historical documents is crucial for preserving the cultural heritage of the society. An important step in this process is converting scanned images to text using Optical Character Recognition (OCR), which can enable further search, information extraction, etc. Unfortunately, this is a hard problem as standard OCR tools are not tailored to deal with historical orthography as well as with challenging layouts. Thus, it is standard to apply an additional text correction step on the OCR output when dealing with such documents.
In this work, we focus on Bulgarian, and we create the first benchmark dataset for evaluating the OCR text correction for historical Bulgarian documents written in the first standardized Bulgarian orthography: the Drinov orthography from the 19th century.
We further develop a method for automatically generating synthetic data in this orthography, as well as in the subsequent Ivanchev orthography, by leveraging vast amounts of contemporary literature Bulgarian texts. We then use state-of-the-art LLMs and encoder-decoder framework which we augment with diagonal attention loss and copy and coverage mechanisms to improve the post-OCR text correction. 
The proposed method reduces the errors introduced during recognition and improves the quality of the documents by 25\%, which is an increase of 16\% compared to the state-of-the-art on the ICDAR 2019 Bulgarian dataset. We release our data and code at \url{https://github.com/angelbeshirov/post-ocr-text-correction}.}


\maketitle

\section{Introduction}\label{intro}

\renewcommand*{\thefootnote}{\fnsymbol{footnote}}
\footnotetext[3]{Work done while Momchil was in the Sofia University, prior to joining Amazon. }
\setcounter{footnote}{0}
\renewcommand*{\thefootnote}{\arabic{footnote}}

In the last years, many libraries and museums have made a step towards datafication, trying to make their documents and papers fully accessible, searchable and processable in digital form. Having this information in digital form helps with preserving the cultural heritage of the society. The documents are digitized using Optical Character Recognition (OCR) which is the process of identifying typed or handwritten printed characters within a document and converting them automatically into computer codes which can then be processed as textual data. 

However, for documents written in old Bulgarian spelling, the process often introduces errors in recognition because of the challenging layouts and orthographic variety in the documents due to the nine language reforms applied to the Bulgarian language prior to 1945. As a result of these errors, other applications like Named Entity Recognition, Part-of-speech tagging, text summarization and others are negatively impacted. According to \cite{noisyner} if the word error rate of the digitized text increases from 0\% to 2.7\% the F1-score of the NER decreases by 3\%.
Nowadays, most modern OCR engines manage to recognize around 99\% of the characters in high-quality documents. For an average word length of 5, this means that one out of every twenty words has an error. For Bulgarian historical documents, the percentage of correctly recognized characters is lower as some of them are not used anymore in the current spelling.

The large amount of transcribed data required to train an OCR model from scratch is not available for the resource-poor Bulgarian historical spellings. Instead, our focus is on post-correcting the produced output and we show that using this approach can significantly improve the quality of the raw OCRed text. Additionally, the characteristics of the Bulgarian language are similar to other Slavic languages, so the methods presented in this paper can easily be adapted for them as well.

We have created a benchmark dataset for the first standardized Bulgarian orthography: the Drinov orthography from the 19th century. For the subsequent Ivanchev orthography, we use the dataset provided by the ICDAR 2019 competition \cite{icdar2019}. Our proposed method reduces the errors introduced during recognition and improves the quality of the documents by 25\%, which is an increase of 16\% compared to the best contender from the ICDAR 2019 on the Bulgarian dataset. Finally, due to scarce training data for these historical Bulgarian spellings, we develop a method which can convert modern Bulgarian spelling into a historical one thus allowing us to leverage the large amount of contemporary literature Bulgarian texts. The method was used in order to generate synthetically noised samples by using the confusion matrix of the isolated misspellings from each dataset. We have released our data and code on GitHub\footnote{\url{https://github.com/angelbeshirov/post-ocr-text-correction}}.

In summary, our contributions are the following:
\begin{enumerate}
  \item A new benchmark dataset for post-OCR text correction on the first standardized Bulgarian orthography: the Drinov orthography from the 19th century. This is the first dataset that uses this orthography.
  \item A method for synthetic data generation, which automatically transforms modern Bulgarian spelling into a resource-poor historical spelling using concrete spelling and grammatical norms. 
  \item A novel post-OCR text correction method that utilizes pretrained LLMs and encoder-decoder framework augmented with diagonal attention loss, copy and coverage mechanisms.
\end{enumerate}

\section{Related Work}\label{relwork}

Post-OCR text correction has been an important problem ever since the inception of OCR technology in the 1960s. According to \cite{methodspostocr} there are three main approaches in order to improve the quality of the OCR output: changing the input images, optimizing the OCR system and post-processing the output. In the current research, we focus on the last option. 

The post-OCR text correction process consists of two interdependent tasks: error detection and error correction. Additionally, there are two types of errors that the OCR model could make: real-word and non-word errors \cite{error_types}. If a token is not a lexicon entry, it is considered a non-word error and if a valid word occurs in a wrong context, then it is a real-word error. The real-word errors are harder to detect and correct since this can be done only by leveraging the context around the word. For example, consider the phrases ``glowing brightly'' and ``growing bnghtly'', a non-word error is ``bnghtly'' and ``growing'' is a real-word error.

In the beginning, the first approaches were based on dictionaries \cite{dict1}, \cite{dict2}, \cite{dict3}. The error detection task consists of checking if a word is present or not in a predefined dictionary. For the error correction, the most similar candidates from the dictionary are computed using similarity measures like Levenshtein distance \cite{levenshtein}, longest common sub-sequence or confusion weight. The International Conference on Document Analysis and Recognition (ICDAR) organized two competitions in 2017 and 2019 for correcting mistakes in digitized documents \cite{icdar2017} \cite{icdar2019}. In the first competition, the best-performing model \cite{besticdar2017} was based on machine translation. They use an ensemble of statistical and neural machine translation models on the character level. In the second competition in 2019, the winner was Clova AI team. They fine-tuned a pretrained BERT model \cite{devlin-etal-2019-bert} with convolutional and fully-connected layers to identify OCR errors. Afterwards, a character-level sequence-to-sequence model was used to correct the erroneous tokens.

Since then, the interest in applying pre-trained language models for post-OCR text correction has increased.  Embeddings from BERT are used in \cite{nguyen_bert} for detecting errors, after which a method based on machine translation corrects them. Another approach is in \cite{zhang-etal-2020-spelling} where again BERT embeddings are used to detect errors after which a BERT model is fine-tuned on MLM to correct these mistakes. In \cite{synthetic} a sequence-to-sequence correction model was trained to correct a synthetically corrupted corpus, constructed by injecting uniformly distributed insertion, deletion and substitution errors into the dataset. The use of synthetic dataset led to an overall improvement. The above studies clearly demonstrate the potential of the pre-trained language models for post-OCR text correction.

\section{Methodology}\label{methodology}
In this section, we define the problem and the pipeline of post-OCR text correction. Afterwards, we describe the error detection and correction in details with the models which we have used to tackle both of these problems.

\subsection{Problem formulation}\label{probform}
The task of post-OCR text correction is defined formally as follows: given a sequence of characters \(C = c_1c_2c_3 \cdots c_n\), the goal is to find the correct sequence of characters \(W = w_1w_2w_3 \cdots w_m\), which matches with the original printed text. Note that it is possible \(n \neq m\) as some symbols might have been missed or a word ending might not have been recognized correctly by the OCR engine. If \(C = W\), then we will consider the word to be correct, otherwise it is incorrect. 

There are two types of errors that the OCR engine can make - real-word and non-word errors. If \(C \not\in D\) where \(D\) is some dictionary we will say that the error is of type non-word error. If  \(C \in D\) then we will say that the error is a real-word error. 

\begin{figure*}
    \centering
	\includegraphics[width=1.9\columnwidth]{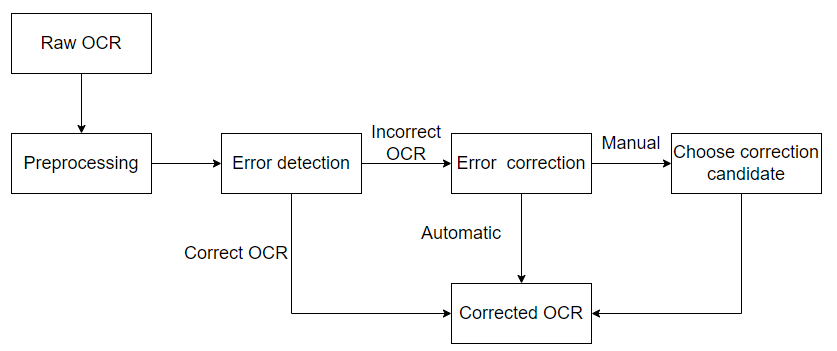}
	\caption{
	    Pipeline for post-OCR text correction.
	}
    \label{fig:pipeline} 
\end{figure*}

The pipeline for the post-OCR text correction process is shown in Figure~\ref{fig:pipeline}. The first step in the pipeline is preprocessing, where the data is tokenized, cleaned and prepared to be fed to the next error detection step. The error detection classifies each token as correct or incorrect. Afterwards, the classified as erroneous tokens are fed into the error corrector model which either directly corrects them or suggests a list of candidates if the text correction is executed in a semi-automatic setting. Splitting the task to error detection and correction adds the benefit of increasing the proportion of OCR errors which are fed into the corrector model and decreases the amount of erroneously changed correct tokens since the tokens classified as correct are not passed through the error corrector model.

\subsection{Error Detection} \label{sec:ed}

The goal of the error detection task is to determine whether a token was correctly recognized by the OCR engine or not. In other words, it is a binary classification. In the text correction pipeline shown in Figure~\ref{fig:pipeline}, only the erroneous tokens will be passed downstream to the error correction task. This is the reason why we are interested in minimizing the incorrectly classified tokens as erroneous, i.e. false positives. Therefore, for the error detection task precision is preferred over recall.

We have used two naive baselines for the error detection task. The first one is based on the CLaDA-BG dictionary \cite{clada} which contains over 1.1 million unique words in all of their correct forms. During inference, if the token is present in the dictionary, then it is considered to be correct, otherwise, it is classified as incorrect. The second baseline uses fastText \cite{fastText}, which is a library for the classification and learning of word representations. As the model is trained on subword ngrams, it can handle out-of-vocabulary words, unlike the CLaDA-BG dictionary. The data is transformed into the format $ \textunderscore\textunderscore label \textunderscore\textunderscore<0|1>\; \; <word> $, where 0 is the label for correct words and 1 for incorrect. 

Pre-trained language models are very effective and achieve state-of-the-art results in solving different NLP tasks, including token classification. We experimented with various language models for solving the task in a context-dependent manner. The best-performing model is the multilingual DeBERTaV3 \cite{he2021deberta}, which improves the BERT \cite{devlin-etal-2019-bert} and RoBERTa \cite{roberta} models using disentangled attention and enhanced mask decoder. The model was trained using the CC100 data, containing 100+ languages including Bulgarian.

For all pre-trained language models, the process of fine-tuning is very similar and consists of the following steps. Initially, we get the tokenized data from the preprocessing step aligned at the token level and we apply the specific model tokenization, which can be WordPiece \cite{wordpiece}, BPE \cite{bpe} or SentencePiece \cite{sentencePiece} for different language models. When a token is split into multiple sub-tokens, each of the sub-tokens receives the label of the token, i.e. if a token is erroneous, then all of its sub-tokens will also be labelled as erroneous. The embeddings produced are then fed as inputs to the token classification language model, which consists of a language model and a fully-connected layer. The produced output contains classified sub-tokens, which have to be merged back into token-level predictions.  This happens by taking every sub-token into consideration and if any of them are predicted to be erroneous, then the whole token is classified as erroneous. The experiments with all language models are shown in Section~\ref{sec:experiments}.

\subsection{Error Correction} \label{sec:ec}

The error correction task consists of transforming each token classified as erroneous from the detector model to the sequence of characters. We use k-nearest neighbors as a naive baseline model. The method consists of finding the nearest neighbors from the CLaDA-BG dictionary \cite{clada} for each erroneous token using Levenshtein distance as a proximity measure. We only consider distances of one and two. If multiple tokens are found with the same distance, then word frequency is used to pick the best one and if no candidates are found, then the token remains unchanged.

As a second approach, we employ neural machine translation at character level to translate the classified as erroneous tokens into their corrected versions. Our model is a character-level sequence-to-sequence model with attention. The encoder is a bidirectional LSTM \cite{lstm} and transforms the characters into a sequence of hidden states $\{h_i\}$. During the decoding process, the model uses an attention mechanism in order to decide what information to use from the encoder's hidden states. The embeddings between the encoder and the decoder are shared and cross-entropy loss between the predicted output and the gold standard text is used as an initial loss function. 

Additionally, some enhancements are implemented to the base sequence-to-sequence model, similar to the ones proposed in \cite{rijhwani-etal-2020-ocr}. The first enhancement is diagonal attention loss. According to \cite{error_types}, the most frequent OCR errors are substitution errors, which means that the OCR engine confused a character with another one. This is expected, as the OCR is most likely to confuse a character rather than hallucinate or completely ignore pixels from the original document. The nature of the OCR is such that, errors are happening for characters that look similar. This is not the case for example in regular spell checking where common mistakes are for characters which are next to each other on the keyboard or the order of the characters is reversed due to fast typing. Therefore, during correction, reordering of characters almost never occurs, so we expect the attention weights of the model to have higher values close to the diagonal of the encoder. The diagonal loss $L_{\mathrm{diag}}$ is added to the final model loss so that the elements from the attention vector that are not within \textit{m} steps from the current timestep \textit{t} are added to the loss. In this way, we encourage elements away from the diagonal to have lower attention scores.
\[
    L_{\mathrm{diag}} = \sum_t \left( \sum_{i=1}^{k-m}\alpha_{t,i} + \sum_{i=t+m}^N\alpha_{t,i} \right)
\]

The second enhancement is a copy mechanism, which enables the model to copy characters from the source directly rather than generating a character on each step. This allows us to leverage the overlap of correctly recognized characters between the raw OCRed text and the gold standard output. We use the copy mechanism first proposed in \cite{see}. It calculates a \textit{generation probability} $P_g$ for each time step \textit{t} which is used as a soft switch to choose whether to generate a character or copy directly from the source. Copying a character from the input sequence happens with probability $1 - P_g$ by sampling the encoder attention distribution. For time step \textit{t}, the \textit{generation probability} is calculated from the context vector $c_t$, the decoder state $s_t$ and the decoder input $x_t$:
\[
    P_g = \sigma(w_c^\intercal c_t + w_s^\intercal s_t + w_x^\intercal x_t + b )
\]

The final enhancement is the coverage mechanism, which keeps track of the attention distribution over all past time steps in a coverage vector: 
\[ 
    g_t = \sum_{i=0}^{t-1} \alpha_i 
\]

Intuitively, the coverage vector is a distribution over the erroneous token's characters, which represents how much coverage each character has received.
Additionally, we also add it as a separate parameter when calculating the attention distribution $ \alpha_t$:
\[ 
    e_{t,i} = \mathrm{vtanh}(W_1S_{t-1} + W_2h_i + W_gg_{t,i}) 
\]
\[ 
    \alpha_t = \mathrm{softmax}(e_{t,i}) 
\] 

The coverage vector is also used to encourage the model to attend to different locations by adding a coverage loss $L_c$:
\[
    L_c = \sum_t\sum_{i=0}^N \min(\alpha_{t,i}, g_{t,i}) .
\]

The final loss of the model is calculated as:
\[
    L = L_{ce} + L_{\mathrm{diag}} + L_c .
\]

In Section~\ref{sec:experiments} we have evaluated the combinations of these enhancements and we see significant improvement when all of them are applied at the same time in the final sequence-to-sequence model.

\begin{figure*}[t!]
    \centering
	\includegraphics[width=1.9\columnwidth]{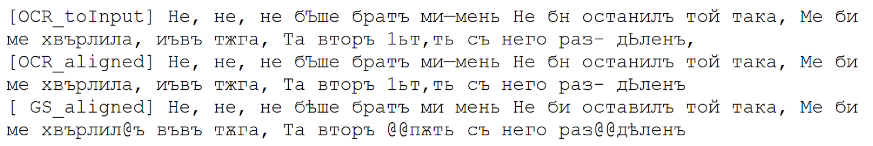}
	\caption{
	    Example sentence from the ICDAR 2019 dataset.
	}
    \label{fig:sample_sentence} 
\end{figure*}

\begin{figure*}[t!]
    \centering
	\includegraphics[width=2\columnwidth]{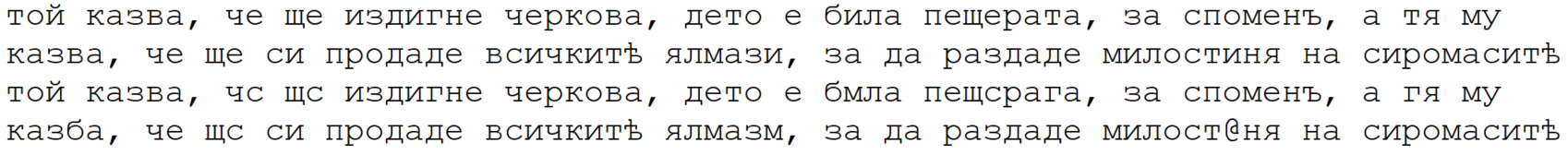}
	\caption{
	    Example of a synthetically generated sentence pair in the Ivanchev orthography. The first sentence is correct, while the second is misspelled.
	}
    \label{fig:sample_sentence_syn} 
\end{figure*}

\section{Data}

This section describes the datasets used for training and evaluating the models as well as the process of generating the synthetic dataset.

\subsection{Datasets}

There is a substantial orthographic variety in the Bulgarian historical documents due to the nine language reforms applied to the Bulgarian language. In 1945 the current Bulgarian orthography was accepted and prior to that, there were two major orthographies that were used - the Drinov orthography and the Ivanchev orthography. 

We have used the Bulgarian dataset provided by the ICDAR 2019 competition on post-OCR text correction. The dataset contains around 200 documents written in the Ivanchev orthography.
The data comes aligned at the character level, but we needed it at the token level. To get that alignment, we tokenize the gold standard based on white space and for each token, we take the same number of characters from the aligned OCRed version. By splitting the data this way we avoid a situation where the incorrectly split OCRed word also splits the GS words.  After that, we remove the special alignment symbols: ``@'' and ``\#'', and the resulting tokens are compared in order to set the labels as \textit{0} for correct and \textit{1} for incorrect. An example sentence from the ICDAR 2019 dataset is shown in Figure~\ref{fig:sample_sentence}.

Since the ICDAR 2019 dataset contains documents only in the Ivanchev orthography we release a small benchmark dataset, which contains documents in the Drinov orthography - the Drinov Orthography for Post-OCR Correction dataset. The corpus was created by annotating a historical newspaper collection provided to us by the National Library ``Ivan Vazov'' (NLIV) in Plovdiv, Bulgaria\footnote{\url{https://digital.libplovdiv.com/en}}. We consider printed versions of these documents, which we manually annotate and align at the character level in the same format as the one from the ICDAR 2019 competition. The number of sentences, words and errors, as well as mean CER $\mu$ and standard deviation CER $\sigma$ of the two datasets are shown in Table~\ref{table:stats}.

\begin{table*}
\centering
\caption{Sentence, word, and error quantities, mean CER and standard deviation CER.}
\setlength{\extrarowheight}{2pt}
\begin{tabular}{p{0.2\textwidth}>{\raggedleft}p{0.13\textwidth}>{\raggedleft}p{0.13\textwidth}>{\raggedleft}p{0.13\textwidth}>{\raggedleft}p{0.13\textwidth}>{\raggedleft\arraybackslash}p{0.13\textwidth}}
\toprule
    Dataset & \# Sentences & \# Words & \# Errors & $\mu$ CER & $\sigma$ CER \\
    \hline
    ICDAR 2019 & 4,900 & 68,511 & 25,703 & 16.65 & 16.13 \\
    DOPOC & 227 & 5,152 & 589 & 2.94 & 1.67 \\
    \hline
\end{tabular}
\label{table:stats}
\end{table*}

\begin{figure}
    \centering
    \begin{minipage}{0.5\textwidth}
        \centering
        \includegraphics[width=0.9\textwidth]{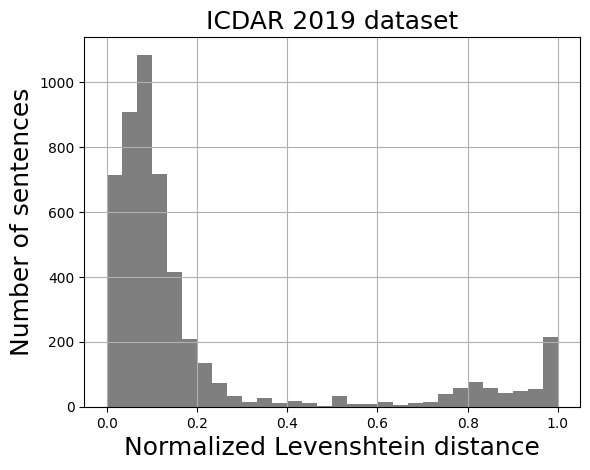}
    \end{minipage}\hfill
    \begin{minipage}{0.5\textwidth}
        \centering
        \includegraphics[width=0.9\textwidth]{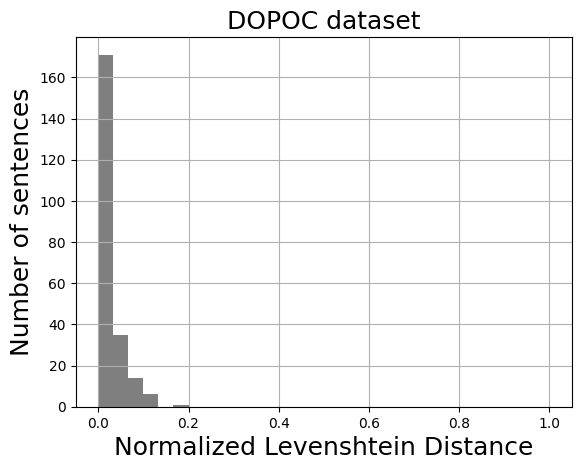}
    \end{minipage}
    \caption{Distribution of the normalized Levenshtein distance for the ICDAR 2019 and DOPOC datasets.}
    \label{fig:norm_lev_trainicdar} 
\end{figure}

However, some sentences are too noisy for learning, as they either contain too many typos, unrecognized or misaligned words. We discard those during training and evaluation by using the normalized Levenshtein distance between the OCRed sentence and its corresponding gold standard. We only consider sentences having normalized Levenshtein distance less than 0.5.
The distribution of the Levenshtein distances from the two datasets are shown in Figure~\ref{fig:norm_lev_trainicdar}. 

\subsection{Synthetic Data} \label{sec:sd}
Since the available data is scarce for the Bulgarian historical spellings, we generate a large amount of synthetically constructed misspelled-correct sentence pairs to support training. We have used around 100k sentences from old Bulgarian books available online in the current orthography. The process of synthetic data generation consists of the following two steps: first, we transform the data from modern to the historical spelling of the respective dataset. In order to do that, we have developed a method which automatically converts from modern to one of the two major historical spellings. The method uses spelling transformation rules \cite{codification} and a morpho-syntactic tagging model for Bulgarian \cite{morph_bg} to transform between the two orthographies. 
For example, the Bulgarian verb ``пея'', which translates to ``sing'' in English, after the conversion is applied results in ``пѣѫ'' and ``пѣя'' in the Drinov and Ivanchev orthographies respectively.
After the word is converted, we insert noise into it by replacing a character with another one using the confusion matrix of each dataset (ICDAR 2019 and DOPOC). The confusion matrix contains the character error frequencies from the documents and is created automatically by tokenizing the OCR aligned and GS aligned sentences and comparing their characters. The amount of generated noise should be similar to the actual noise from real-world OCRed documents, otherwise it will have detrimental impact on the accuracy of the model \cite{genocr}. This is accomplished by the calculated probabilities from the confusion matrix. An example of synthetically generated sentence pair is shown in Figure~\ref{fig:sample_sentence_syn}.

\section{Experiments} \label{sec:experiments}

This section describes the experimental setup and performance of the evaluated models for error detection and correction tasks on both datasets.

\subsection*{Setup}
For evaluating the error detection models we use the standard metrics for classification tasks: precision, recall and F1-score. To evaluate error correction we use \% of improvement, which is calculated by comparing the weighted sum of the Levenshtein distances between the noisy raw OCRed text $ocr$ and the corrected one $c$ with the gold standard $gs$. 
\[
\% \:\: \mathrm{improvement} = \frac{\mathrm{lev}(ocr, gs) - \mathrm{lev}(c, gs)}{\mathrm{len}(gs)}
\]
The train dataset is split further into train and development datasets. The train dataset contains 90\% of the examples and the development dataset contains the rest 10\%. For the pretrained language models, we have used the Huggingface's transformers library \cite{huggingface}. We are fine-tuning them for 10 epochs using AdamW as an optimizer and cross-entropy loss between the predicted output and the gold standard text. The learning rate is \textit{3e-5} and the sequence length is 100. The error correction model was trained for 60 epochs and beam search was used at inference time. We trained all models on Tesla T4 GPU 16GB with early stopping on the development set. The models are trained on the ICDAR 2019 dataset to which we add synthetically generated data for the specific orthography as described in Subsection~\ref{sec:sd}. 

The evaluated language models for error detection are: BERT \cite{devlin-etal-2019-bert}, DistilmBERT \cite{distilbert}, RoBERTa\footnote{https://huggingface.co/iarfmoose/roberta-base-bulgarian} \cite{roberta}, XLM-RoBERTa \cite{xlmroberta}, XLM-MLM \cite{xlmmlm100} and mDeBERTaV3 \cite{he2021deberta}, as well as two naive baselines - CLaDA-BG and fastText. 

\begin{table*}[t]
\centering
\caption{\textbf{Error detection:} evaluation results in terms of P, R, and F1. We present the results with and without synthetic data and highlight the best model for each orthography. $\blacktriangledown$ denotes ICDAR 2019 baseline and $\triangledown$ denotes naive baseline model.
}
\label{table:resed}
\setlength{\extrarowheight}{1.5pt}
\begin{tabular}{m{9em} m{9em} m{2.5em} m{2.5em} m{2.5em} !{\extracolsep{6pt}} m{2.5em} m{2.5em} m{2.5em}}
\toprule
      &  & \multicolumn{3}{c}{\normalsize ICDAR 2019} & \multicolumn{3}{c}{\normalsize DOPOC} \\\cline{3-5} \cline{6-8}
    Model         & Synthetic Data  &   P     &   R   &    F1    &   P   &   R    &   F1     \\[1pt]
    \hline
    CCC$\blacktriangledown$    & No     &   -   &  -  &  0.77  & -  &  -  &  -   \\  
    \hline
    CLaDA-BG$\triangledown$     & No     &   0.77   &  0.55  &  0.64  &  0.28  &  0.56  &  0.37   \\  
                  & Yes    &   0.79   &  0.57  &  0.66  &  0.30  &  0.53  &  0.37   \\  
    \hline
    FastText$\triangledown$     & No     &   0.66   &  0.81  &  0.73  &  0.33  &  0.61  &  0.43   \\  
                  & Yes    &   0.83   &  0.46  &  0.60  &  0.46  &  0.29  &  0.35   \\ 
    \hline
    mBERT         & No     &   0.79   &  0.80  &  0.81  &  0.68  &  0.26  &  0.38   \\  
                  & Yes    &   0.83   &  0.84  &  0.84  &  0.60  &  0.48  &  0.53   \\ 
    \hline
    DistilmBERT   & No     &   0.78   &  0.80  &  0.80  &  0.68  &  0.24  &  0.36   \\  
                  & Yes    &   0.83   &  0.83  &  0.82  &  0.61  &  0.42  &  0.49   \\ 
    \hline
    RoBERTa       & No     &    0.75  &  0.79  &  0.77  &  0.71  &  0.26  &  0.39   \\  
                  & Yes    &    0.75  &  0.81  &  0.78  &  0.59  &  0.38  &  0.46   \\ 
    \hline
    XLM-RoBERTa   & No     &    0.81  &  0.85  &  0.83  &  0.78  &  0.29  &  0.42   \\  
                  & Yes    &    0.84  &  0.88  &  0.85  &  0.70  &  0.52  &  0.60   \\ 
    \hline
    XLM-MLM-100   & No     &    0.81  &  0.84  &  0.83  &  0.75  &  0.26  &  0.39   \\  
                  & Yes    &    0.83  &  0.84  &  0.84  &  0.61  &  0.47  &  0.53   \\ 
    \hline
    mDeBERTaV3    & No     &    0.87  &  0.89  &  0.88  &  0.78  &  0.33  &  0.46   \\  
                  & Yes    &    0.89  &  0.91  &  \textbf{0.90}  &  0.74  &  0.59  &  \textbf{0.65}   \\ 
    \hline
\end{tabular}
\end{table*}

\begin{table*}
\centering
\caption{\textbf{Error correction:} evaluation results in terms of \% improvement. kNN is k-nearest neighbors with Levenshtein distance, Seq2Seq Base is a base character-level sequence-to-sequence model with attention. Seq2Seq Copy includes a copy mechanism and Seq2Seq Final includes all enhancements - diagonal attention loss, copy and coverage mechanisms. $\blacktriangledown$ denotes ICDAR 2019 baseline and $\triangledown$ denotes naive baseline model.
}
\label{table:resec}
\setlength{\extrarowheight}{1pt}
\begin{tabular}{m{9em} m{9em} m{7em} m{5em} }
\toprule
    Model              &  Synthetic Data   &     ICDAR 2019  &   DOPOC           \\
    \hline
    CCC$\blacktriangledown$    & No                &    9\%             &   -              \\
    \hline
    kNN$\triangledown$ & No                &    -2.4\%          &  -3.2\%         \\
                       & Yes               &    -1.6\%          &  -2.1\%         \\
    \hline
    Seq2Seq Base       & No                &    17.1\%          & 19.3\%           \\
                       & Yes               &    19.8\%          & 22.1\%           \\
    \hline
    Seq2Seq Copy       & No                &    19.2\%          & 21.7\%           \\
                       & Yes               &    21\%            & 25.3\%           \\
    \hline
    Seq2Seq Final      & No                &    22.7\%          & 25.3\%            \\
                       & Yes               &    \textbf{25.4\%} & \textbf{26.9\%}   \\
    \hline
\end{tabular}
\end{table*}

\subsection*{Results}

The results from the experiments are shown in Table~\ref{table:resed}. The best-performing model for both datasets is the multilingual DeBERTaV3 with synthetic data during training. On the ICDAR 2019 competition the best-performing model (CCC) achieves F1-score of 0.77 on the Bulgarian dataset, which we manage to improve by +0.13. DeBERTaV3 gets a 0.65 F1-score on the DOPOC dataset, which is lower than the results from the ICDAR 2019 dataset. We attribute these lower detection scores to two factors: better document quality of the dataset which makes the errors harder to detect and lack of original annotated train dataset in the Drinov orthography. Without synthetic data during training, the model is not able to learn a reasonable distribution of the Drinov orthography. However, after adding the synthetically noised data, we see a significant improvement from 0.46 to 0.65 F1-score for DeBERTaV3. The CCC baseline was not run on the DOPOC dataset since the hyperparameters of the model were not shared in order to properly finetune it on the Drinov orthography. Thus, we only use it as a baseline comparison against the best-performing model from the ICDAR 2019 competition.


In Table~\ref{table:resec} the results for the error correction models described in Subsection~\ref{sec:ec} are shown. The kNN baseline model is actually worsening the quality of the raw OCR because it does not consider the individual mistaken characters and is often picking up candidates which are correct according to the dictionary but in reality, have greater edit distance than the original text. The character-based Seq2Seq model achieves good results compared to the baseline. After adding copy mechanism to the original Seq2Seq model, the performance is notably better for both orthographies. This can be explained since the first pass OCR manages to predict most of the characters correctly. The best-performing model combines all enhancements described in Subsection~\ref{sec:ec} and is trained using both the original and synthetic datasets. The Seq2Seq Final achieves 25.4\% of improvement on the ICDAR 2019 dataset, which improves the results of the best-performing model from the competition (CCC) by +16.4\% of improvement. On the DOPOC dataset, we are able to properly correct most of the erroneous tokens and get a 26.9\% of improvement on the original raw OCRed documents. The above results clearly demonstrate that by leveraging synthetic data we can further improve the quality of digitized historical documents written in any of the two main Bulgarian historical spellings.

\subsection*{Error Analysis} \label{sec:erroranalysis}
\begin{figure*}
\centering
\begin{tabular}{m{9em} m{4em} m{10em} m{5em} m{6em} }
    & \multicolumn{2}{l}{\normalsize Errors fixed by post-correction} & \multicolumn{2}{c}{\normalsize Errors \textbf{not} fixed by post-correction} \\
    & & & & \\
    Raw OCR             &  разб\textcolor{red}{н}тъ   &    испов\textcolor{red}{е}дн\textcolor{red}{п}цит\textcolor{red}{б}  &   с\textcolor{red}{мп}ватъ  &   в\textcolor{red}{п}д\textcolor{red}{ня}ъ           \\
                        &                                  &                                          &                                  &           \\
    GS OCR   & разб\textcolor{blue}{и}тъ    &     испов\textcolor{blue}{ѣ}дн\textcolor{blue}{и}цит\textcolor{blue}{ѣ}  &    с\textcolor{blue}{тѫп}ватъ &    в д\textcolor{blue}{им}ъ           \\
                        &                                  &                                          &                                   &          \\
    Post-corrected      & разб\textcolor{green}{и}тъ    &   испов\textcolor{green}{ѣ}дн\textcolor{green}{и}цит\textcolor{green}{ѣ}    &    с\textcolor{red}{ми}ватъ   &   в\textcolor{red}{и}д\textcolor{red}{ня}         \\
    & & & \\
\end{tabular}
\caption{Our model manages to fix most of the errors, as shown in the examples from the first two columns. However, in some cases it fails to properly correct the error, as demonstrated in the two subsequent columns.}
\label{fig:erroranalysis}
\end{figure*}

\begin{table}[t]
\centering
\caption{OCR error types with examples.}
\setlength{\extrarowheight}{2pt}
\begin{tabular}{p{0.2\textwidth}>{\raggedright\arraybackslash}p{0.22\textwidth}}
\toprule
    Error Type & Example \\
    \hline
    Misrecognized character & more $\xrightarrow{}$ moro   \\
    Missing character & there $\xrightarrow{}$ here \\
    Hallucination & where $\xrightarrow{}$ wherever \\
    Run-on & he ran $\xrightarrow{}$ heran \\
    Incorrect split & locomotive $\xrightarrow{}$ loco motive \\
    \hline
\end{tabular}
\label{table:error_examples}
\end{table}

In Table~\ref{table:error_examples} we have shown the types of errors that an OCR engine makes with examples in English. Our models are able to detect and correct most of the errors introduced during the OCR process. Examples of errors fixed by the post-OCR correction method are shown in Figure~\ref{fig:erroranalysis}. The model is also able to successfully correct multiple incorrectly recognized characters.

However, we have noticed that the models do not perform well when it comes to fixing word segmentation errors, which are incorrectly recognized word endings. These types of errors are divided into two types \cite{error_types}: run-on and incorrect split errors. In case of a run-on error, the model usually simply misspells the input word without adding the necessary space, whereas incorrect split errors often result in misspelled words rather than being properly combined. An example of an unsuccessful attempt to fix a run-on error is shown in the rightmost column in Figure~\ref{fig:erroranalysis}. The counts of word segmentation errors from our datasets are shown in Table~\ref{table:seg_error_stats}.

\begin{table}[t]
\centering
\caption{Quantities of word segmentation errors in the ICDAR 2019 and DOPOC datasets.}
\setlength{\extrarowheight}{2pt}
\begin{tabular}{p{0.2\textwidth}>{\raggedleft}p{0.1\textwidth}>{\raggedleft\arraybackslash}p{0.1\textwidth}}
\toprule
    \# errors & ICDAR 2019 & DOPOC \\
    \hline
    Total & 25703 & 589 \\
    Word segmentation & 3503 & 44 \\
    Run-on & 2423 & 11 \\
    Incorrect split & 1080 & 33 \\
    \hline
\end{tabular}
\label{table:seg_error_stats}
\end{table}

There is some research done on specifically fixing word segmentation errors, in \cite{wordseg} a character-level Seq2Seq model is utilized for correcting these types of errors separately. The training and test data are scientific texts from the electronic submissions of the ACL collection. Word-level data is first transformed into character-level by removing all white spaces between words in the input data and replacing them with ``\#\#''. The model is then trained on this data to learn the word boundaries represented by the special symbols ``\#\#'' and use it to correct word boundary errors in the test dataset with achieved precision and recall of above 0.95.

\section{Conclusion and Future Work}
In this paper, we have presented a new post-OCR text correction method for Bulgarian historical documents. Our proposed methods reduce the errors introduced during recognition and notably improve the quality of the documents. The characteristics of the Bulgarian language are very similar to other Slavic languages, so the methods can easily be adapted for them as well with minimal changes required. 

For solving the error detection task we have used various pre-trained LLMs. The best performing one is DeBERTaV3 which achieves 0.90 F1-score on the ICDAR 2019 dataset and 0.65 F1-score on the DOPOC dataset. Character-level sequence-to-sequence model augmented with diagonal attention loss, copy and coverage mechanisms was used for tackling the error correction task which improves the text quality of the ICDAR dataset by 25\% and of the DOPOC dataset by 27\%. Our results show that we get an improvement over the best performing model in the ICDAR 2019 competition \cite{icdar2019} with +0.13 F1-score for error detection and +16\% for the error correction task. Additionally, we develop a method for generating noised synthetic data, which automatically transforms modern Bulgarian spelling into a historical one by using concrete spelling and grammatical norms which have an outcome in the spelling. This allows us to utilize the large amount of Bulgarian literature text resources available online and adapt them for any of the two main historical spellings.

In future work, we plan to address the word segmentation errors as mentioned in Section~\ref{sec:erroranalysis} the models do not perform very well when it comes to correcting this type of errors. 

\bibliographystyle{unsrt}
\bibliography{main}

\end{document}